\documentclass{article} %
\usepackage[preprint]{goodfire}

\usepackage{microtype}
\usepackage{hyperref}
\usepackage{url}
\usepackage{booktabs}
\usepackage[table]{xcolor}
\usepackage{amsmath}
\usepackage{cleveref}
\usepackage{macros}

\usepackage{etoolbox}
\usepackage{lineno}
\usepackage[color=pink!20,textsize=scriptsize]{todonotes}

\definecolor{rankfirst}{RGB}{158,202,225}   %
\definecolor{ranksecond}{RGB}{198,219,239}  %
\definecolor{rankthird}{RGB}{230,240,250}   %
\newcommand{\rA}[1]{\cellcolor{rankfirst}#1}
\newcommand{\rB}[1]{\cellcolor{ranksecond}#1}
\newcommand{\rC}[1]{\cellcolor{rankthird}#1}

\hypersetup{colorlinks=true, citecolor=primary, linkcolor=primary, urlcolor=primary}

\providecommand{\aut}[1]{\textbf{#1}}
\providecommand{\af}[1]{{\small #1}}

\providecommand{\afn}[1]{\textcolor{antique}{$^{#1}$}}

\newcommand{\costar}{{\color{burgundy}\boldsymbol{\star}}}
\newcommand{\colead}{{\color{burgundy}\boldsymbol{\dagger}}}

\newcommand{\goodfireaff}{%
  \includegraphics[height=18pt]{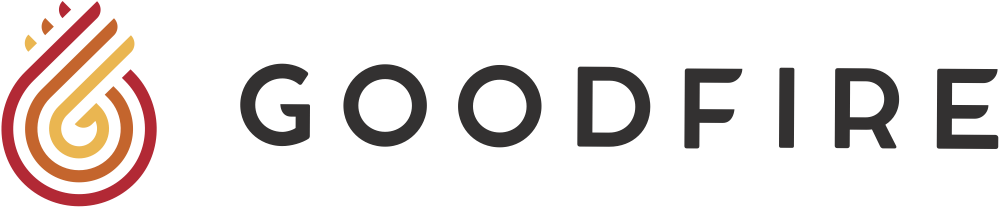} 
}

\newcommand{\authstrut}{\rule[-0.6ex]{0pt}{3.2ex}}
\newcommand{\authorentry}[2]{\authstrut\aut{#1}\afn{#2}}
\newcommand{\afflabel}[2]{\afn{#1}\af{#2}}
\newcommand{\authorsep}{\quad}
\newcommand{\repolink}[1]{%
    {\small
      \href{#1}{\raisebox{-2.8pt}{\includegraphics[height=10pt]{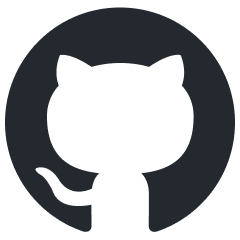}}%
    \textcolor{secondary}{~\nolinkurl{#1}}}
    }%
  }

\newtoggle{goodfireonly}
\togglefalse{goodfireonly}  %
\newcommand{\extline}[1]{\iftoggle{goodfireonly}{}{#1}}
\newcommand{\extaff}[1]{\iftoggle{goodfireonly}{}{#1}}

\newcommand{\paperauthors}{

\authorentry{Aiden Swann}{\costar,\bullet,\extaff{a}} \authorsep
\authorentry{Mathilde Papillon}{\costar,\bullet,\extaff{b}} \authorsep
\authorentry{Mozes Jacobs}{\costar,\bullet,\extaff{c}}  \\[2pt]

\authorentry{Siddharth Boppana}{\bullet} \authorsep
\authorentry{Tal Haklay}{\bullet} \authorsep 
\authorentry{Eric Bigelow}{\bullet} \authorsep
\authorentry{Andrew Lee}{\bullet} \authorsep

\authorentry{Conor Watts}{\bullet} \\[2pt]
\authorentry{Thomas Icard}{\bullet,\extaff{a}} \authorsep 
\authorentry{Mac Schwager}{\extaff{a}} \authorsep
\authorentry{Owen Lewis}{\bullet} \authorsep

\authorentry{Jack Merullo}{\bullet} \\[2pt]
\authorentry{Ekdeep Singh Lubana}{\bullet} \authorsep 
\authorentry{Atticus Geiger}{\colead, \bullet} \authorsep 

\authorentry{Thomas Fel}{\colead,\bullet} \authorsep
\authorentry{Matthew Kowal}{\colead,\bullet} \\[4pt]

$^\costar$\af{Core contributors} \authorsep
\vspace{2mm}
$^\colead$\af{Equal senior contribution} \\
\goodfireaff \\
\vspace{3mm}
\repolink{https://github.com/goodfire-ai/robotics_causal_analysis}
\\
\vspace{1mm}
\extline{%
  \afflabel{\bullet}{Goodfire}\authorsep
  \afflabel{a}{Stanford University}\authorsep
  \afflabel{b}{University of California, Santa Barbara}\authorsep
  \afflabel{c}{Harvard University}\authorsep
}
\vspace{-4mm}
}

\author{\paperauthors}

\title{Causal Analysis of Robot Foundation Models Enables Efficiency Gains}

\begin{document}
\vspace*{-22.5mm}

\maketitle

\vspace{-6mm}
\begin{figure}[h]
    \centering
    \includegraphics[width=1.0\linewidth]{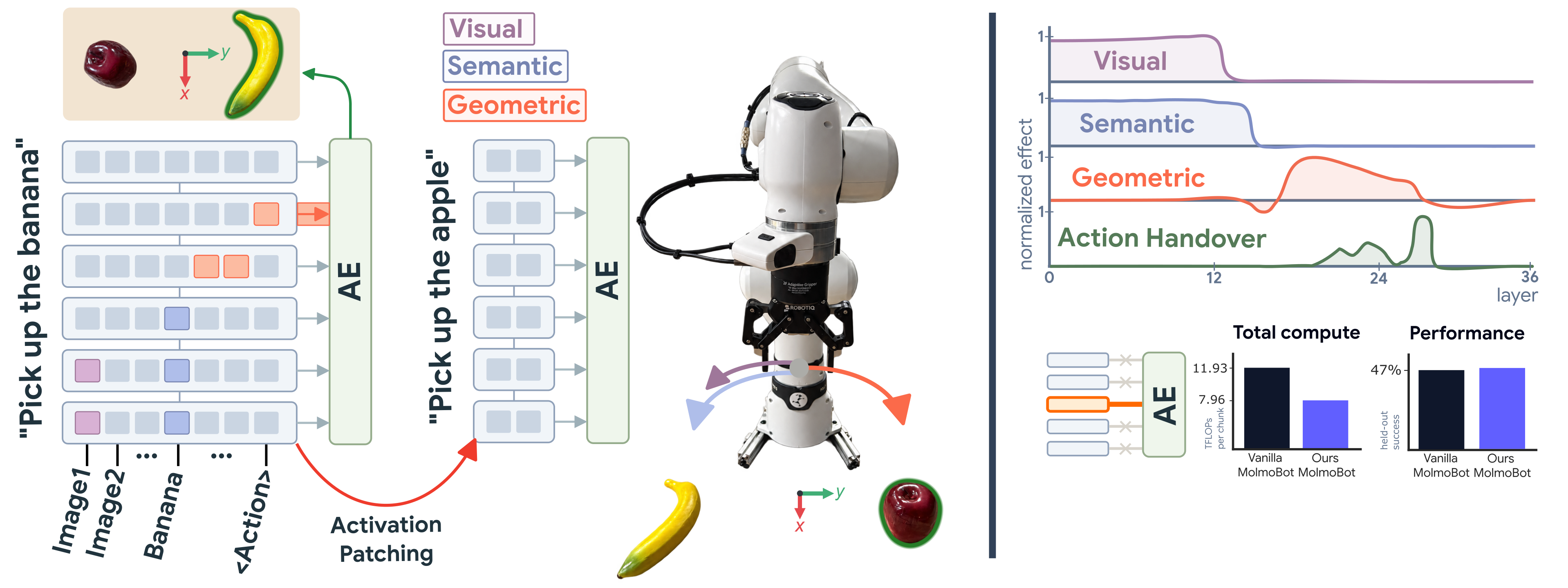}
    \caption{\textbf{Activation patching reveals how a vision-language-action (VLA) model turns perception into action.} (\textbf{Left}) We copy activations from a donor episode (``Pick up the banana'') into a target episode (``Pick up the apple'') with the position of the objects swapped. Visual information (purple) sits in early image tokens, semantic information (blue) in the early noun token, and geometric information (orange) in later text tokens.  The action expert (AE) reads from a narrow band of later layer. (\textbf{Top right}) Schematic of these effects by layer. The action expert reads from a narrow band of layers.(\textbf{Bottom right}) Keeping only this interface in MolmoBot, \textbf{3 of 36 connections}, cuts compute by 33\% with comparable average success (46.7\% vs.\ 46.9\%) over 3000 trials.}
    \label{fig:headline}
\end{figure}

\begin{abstract}

Vision-language-action (VLA) models are a promising approach to general purpose robot manipulation. While the internal computation of vision-language models, which compose the backbone of VLAs, has been studied extensively, less is known about how their perceptual and semantic representations are transformed into action in VLA models. We study this question mechanistically in three frontier VLA models ($\pi_{0.5}$, MolmoAct2, and MolmoBot) by analyzing the causal pathways through which the input image and the instruction shape actions. Across the three models, activation patching shows that causally relevant information influences the action expert (AE) through a narrow band of mid-to-late cross-attention layers, which we term the \emph{action-conditioning interface}. The content of this information changes with token and layer within the model. Early-to-mid layer representations of the target noun token determine which object is selected, whereas mid-to-late layers carry the spatial information in text tokens that redirects the movement when patched. Where spatial information is stored in prompt tokens various based on the architecture. In MolmoAct2 the model forms task-conditioned representations at the text positions that follow the instruction, whereas $\pi_{0.5}$ additionally stores this information at image token positions. We show that these findings can be translated directly into improving model efficiency. We find that restricting MolmoBot's action expert conditioning to only 3 out of 36 layers reduces inference FLOPs by up to $33\%$ while maintaining comparable task success with the default model across 3,000 evaluation episodes (46.7\% vs. 46.9\%), and the same principle carries over to real-robot deployment of $\pi_{0.5}$.
\end{abstract}

\section{Introduction}
\label{sec:intro}

Vision-language-action (VLA) models are a promising route to achieve general-purpose robot manipulation. By adapting large pretrained vision-language models (VLMs) to action \citep{driess2023palm, beyer2024paligemma, deitke2025molmo}, VLAs combine broad visual and semantic knowledge with continuous control and increasingly handle manipulation tasks across objects, environments, and robot embodiments \citep{brohan2022rt,brohan2023rt2,kim2024openvla,black2024pi0,intelligence2025pi05}.

Architecturally, VLAs typically combine a pretrained VLM backbone \citep{chi2023diffusionpolicy,black2024pi0,intelligence2025pi05, bjorck2025groot, deshpande2026molmobot, fang2026molmoact2, wen2025dexvla, wen2025tinyvla} with a separate diffusion or flow-matching action expert that produces continuous control. The two modules are typically joined through a dense, layer-wise interface that conditions the expert on the backbone at every layer. Despite their growing real-world efficacy and partially due to the multi-modal architecture design, these models remain black boxes. VLAs transform thousands of image, language, and proprioceptive tokens through billions of parameters before producing a small set of continuous action tokens, leaving the internal mapping from perception to action an open area of study.

We study three frontier open-source VLAs: $\pi_{0.5}$, MolmoBot, and MolmoAct2 \citep{intelligence2025pi05,deshpande2026molmobot,fang2026molmoact2}. Across all three models, we find that perception and action communicate through a surprisingly narrow channel (1-3 cross-attention layers). Causal influence concentrates in a middle-to-late band of VLM layers, which we term the \emph{action-conditioning interface}. This suggests that the action expert ingests representations which already encode task-relevant spatial information, rather than depending equally on every stage of backbone computation.

We apply our findings of the narrow \textit{action-conditioning interface} between the AE and VLM to improve VLA efficiency. Many prior works have attempted to reduce VLA computation, by skipping or exiting backbone layers \citep{yue2024deer,zhang2025mole,zhang2026a1}, pruning redundant layers or parameters \citep{yang2025efficientvla,nguyen2026fewerlayers,sun2026dropthenrecovery,zhang2026redundancy}, or conditioning the action head on only a subset of backbone layers \citep{bjorck2025groot,shukor2025smolvla}. These methods decide what to remove using heuristics such as representational similarity or benchmark success. We offer a more mechanistic account of why these designs work. Our results show that the representations necessary for action are stored sparsely across tokens and layers.

In this work we show that by leveraging causal analysis we can choose exactly the relevant areas of the model to prune. Restricting conditioning to the causally important layers removes $92\%$ of MolmoBot's cross-attention parameters and reduces the estimated VLM--action-expert floating-point operation count by up to $33\%$ with a similar change in real world measured latency. The simplified architecture achieves task success comparable to the vanilla policy across 3,000 evaluation episodes. The same approach carries over to preliminary trials on a real robot, where sparse and dense $\pi_{0.5}$ policies achieve comparable success rates (67 and 64 of 100 trials). We view these results as a step toward \emph{intentional design} \citep{mcgrath2026intentional} of VLA models, where the internal mechanisms connecting perception to control directly inform how robot policies are built and trained.

Beyond just leveraging the action-conditioning interface we seek to understand and explain it. There are many promising approachs to understanding VLA computation via mechanistic interpretability \citep{grant2026features, bhardwaj2026decoding, molinari2025emergent, khan2025controlling, goyal2026hitchhiker, yang2026interpretable}. In particular, \textit{activation patching} \citep{vig2020causal, geiger2021causal, zhang2023bestpractices, heimersheim2024patching} approaches aim to measure the \textit{causal} role of an internal representation by replacing the target representation from one rollout with a donor from another rollout, while measuring the resulting change in behavior. In language models, such interventions have localized the computations behind factual recall and uncovered computations underlying numerous tasks \citep{meng2022rome,wang2022ioi,geiger2024causalabstractiontheoreticalfoundation,mueller-etal-2026-quest}, helping elucidate the causal relationships between internal model representations and behavior.

Recent work has begun to apply such tools to VLAs, identifying internal directions that steer movement, recovering sparse features tied to semantic concepts, and manipulating observable features to control behavior \citep{haon2025steering,swann2026sae,buurmeijer2026observing}. Closest to our work, \cite{grant2026features} performed activation-injection experiments and showed that patching representations into a VLA can redirect movement toward the spatial location of a donor task. \citet{shi2026vlatrace} used attention knockout \citep{geva2023dissecting}, which blocks attention from selected token positions (such as all image or all language tokens) at chosen layers and measures how the output changes, to characterize which modalities and layers action generation depends on. Together, these results establish that individual VLA representations can be causally manipulated. However, they do not address how task-relevant information is composed across the depth of the backbone and how it reaches the action expert.

We answer these questions with counterfactual comparisons that test visual, semantic, and geometric hypotheses at the interface and within the backbone. We trace how the representations read by the expert reveals a progression across depth, from transferable object selection to spatially grounded action information. The two kinds of information also reside on different tokens: semantic content is carried by the noun token of the instruction, while geometric content is spread across the other prompt tokens and, in $\pi_{0.5}$, the image tokens.

Our contributions are as follows:
\begin{enumerate}[leftmargin=*, itemsep=3pt, parsep=0pt, topsep=3pt]
    \item \textbf{Discovery of a narrow action-conditioning interface.} Across $\pi_{0.5}$, MolmoAct2, and MolmoBot, we identify a narrow middle-to-late region through which the backbone conditions action generation. We show that this interface mediates geometric information about the target object.
    \item \textbf{Mechanistically informed architectural improvements.} We use the identified interface to design substantially sparser VLM--action-expert connections, preserving observed closed-loop performance while reducing the estimated combined operation count by up to $33\%$. We validate the simplified architecture through direct training and adaptation in simulation, and demonstrate real-robot deployment with $\pi_{0.5}$.
    \item \textbf{A causal account of the information supplied to the interface.} We build a counterfactual dataset with over 156 counterfactual pairs across sets of household objects. We run over 775,000 activation patching interventions across $\pi_{0.5}$, MolmoAct2, and MolmoBot specifically designed to differentiate representations within the backbone. Visual information influences behavior through early image tokens, semantic information through the early target noun, and geometric information through later text and readout positions. 
\end{enumerate}

\section{Discovering the Action-Conditioning Interface}
\label{sec:methods}

Our goal is to study how task-relevant information within the vision-language backbone is passed to the action expert. We then use the identified causal interface to create policies that improve efficiency without costing performance.

\paragraph{Models and Notation.}
We study MolmoBot, $\pi_{0.5}$, and MolmoAct2 \citep{deshpande2026molmobot,intelligence2025pi05,fang2026molmoact2}. Each model conditions its continuous action expert on representations from all backbone layers using different attention implementations. MolmoBot supplies layer-wise hidden states through cross-attention, MolmoAct2 supplies layer-wise keys and values, and $\pi_{0.5}$ conditions on a shared multimodal prefix. Their attention masks also differ: in MolmoBot and MolmoAct2, image positions cannot attend to subsequent instruction tokens, whereas $\pi_{0.5}$ uses bidirectional attention within the prefix.

Let $x=(I,u,s)$ denote an input comprising image observations $I$, a language instruction $u$, and proprioceptive state $s$. The policy $F$ produces an action sequence $\bm{a}\in\mathbb{R}^{H\times d_a}$,where $H$ is the prediction horizon and $d_a$ is the number of action components per timestep. We write
\begin{equation}
    \bm{a}=F(x;\epsilon)
    =A\bigl(s,C_1(x),\ldots,C_M(x);\epsilon\bigr),
\end{equation}
where $A$ denotes the action expert together with its action-generation procedure, $\epsilon$ denotes generation noise, and $C_j(x)$ is the backbone-derived conditioning supplied to expert block $j\in\{1,\ldots,M\}$. Direct conditioning on the state $s$ is included where used by the architecture (MolmoBot). In our experiments we hold $\epsilon$ fixed across paired runs and drop it below for simplicity. 

Let $z_\ell(x)\in\mathbb{R}^{N\times d_\ell}$ denote the backbone residual-stream activations at layer $\ell$, with $N$ token positions and hidden dimension $d_\ell$, and let $z_{\ell,T}(x)$ denote the tokens at positions $T$. Here, $z_\ell(x)$ denotes residual-stream activations within backbone layer $\ell$, while $C_j(x)$ denotes the conditioning supplied to action-expert block $j$.

\subsection{Interface interventions.}
Activation patching replaces an internal representation in a recipient run with the corresponding representation from a donor run, then measures the change in output \citep{vig2020causal,geiger2021causal,geiger2024causalabstractiontheoreticalfoundation, zhang2023bestpractices, heimersheim2024patching}.

\emph{Interface patching} replaces the conditioning supplied to selected action-expert blocks $J\subseteq\{1,\ldots,M\}$ while keeping the original VLM computation fixed:
\begin{equation}
    \bm{a}^{\mathrm{int}}_{J,T}
    =
    \left.
        F(x^{\mathrm{o}})
    \right|_{
        C_{j,T}\leftarrow C_{j,T}(x^{\mathrm{cf}}),\; j\in J
    }.
    \label{eq:interface_patch}
\end{equation}
This intervention notation indicates that the original computation is run with the shown internal quantities replaced by their counterfactual values. All other conditioning retains its original value. The action expert continues computing from its modified inputs. We use this type of patching to test where the expert reads causally relevant information from the VLM.

\paragraph{Measuring behavioral transfer.}
Using shared action noise $\epsilon$, we compare the original output with the output under an interface or backbone intervention. We reconstruct trajectories from predicted action commands using forward kinematics, $\bm{\tau}=\mathrm{FK}(\bm{a})$. Let $\bm{\tau}_o$, $\bm{\tau}_{\mathrm{patch}}$, and $\bm{\tau}_{\mathrm{ref}}$ denote the original, patched, and reference trajectories, represented as vectors of stacked positions. $\bm{\tau}_{\mathrm{ref}}$ is the trajectory obtained by changing the original instruction to request the other object while retaining the original scene. We quantify behavioral transfer by the normalized trajectory shift.
\begin{equation}
    E=
    \frac{\left\langle
    \bm\tau_{\mathrm{patch}}-\bm\tau_{\mathrm{o}},
    \bm\tau_{\mathrm{ref}}-\bm\tau_{\mathrm{o}}
    \right\rangle}
    {\left\|\bm\tau_{\mathrm{ref}}-\bm\tau_{\mathrm{o}}\right\|_2^2},
    \qquad \bm\tau_{\mathrm{ref}}\ne\bm\tau_{\mathrm{o}}.
    \label{eq:behavioral_transfer}
\end{equation}
Put simply, $E$ is how far the patch moves the recipient's trajectory toward the reference, as a fraction of the full distance between them. $E=0$ means the patch had no effect in that direction, and $E=1$ means it moved the trajectory all the way to the reference.

\subsection{Action Expert - Interface Patching}
\label{sec:interface_patching}

\begin{figure}[h]
    \centering
    \includegraphics[width=0.99\linewidth]{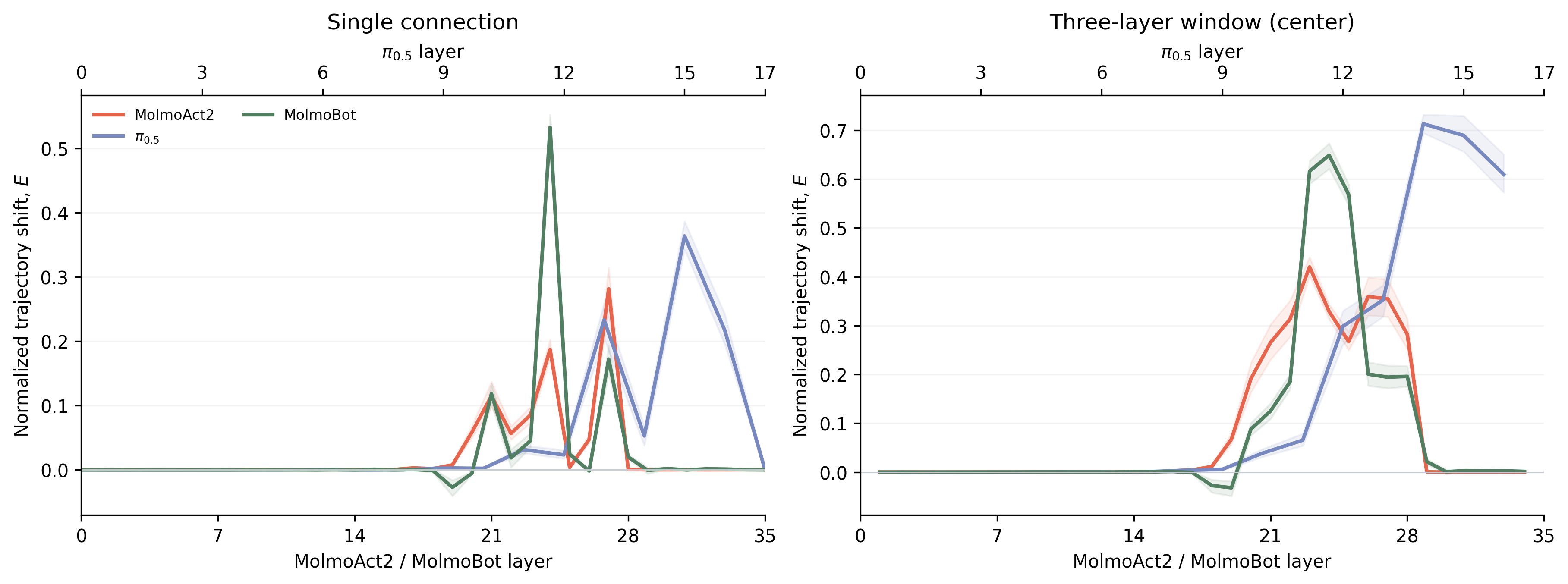}
    \caption{\textbf{Action conditioning concentrates in model-specific layer bands.} Curves show normalized trajectory shift $E$ toward the reference trajectory (same scene other object requested) after patching one action-expert connection (left) or three consecutive connections (right, indexed by the center layer). The results shown are computed across 144,768 patches over 26 scenes via averaging comparisons B and C (\Cref{sec:counterfactuals}).}
    \label{fig:single_sliding_patch}
\end{figure}

We sweep individual connections and three-layer windows using input pairs whose requested targets occupy different locations. We obtain these pairs by changing the requested object in a fixed scene or by exchanging object positions under a fixed instruction (\Cref{sec:counterfactuals}). These experiments use the first timestep of each episode, when both objects are visible. The results are shown in Figure~\ref{fig:single_sliding_patch}.

We observe a highly uneven behavior of influence; e.g., in MolmoBot, patching the action expert at layer 24 alone recovers over $50\%$ of the instructed reach contrast, and patching a three-layer window centered on layer 24 recovers close to $70\%$. Most other individual layers have comparatively little influence (more details in \Cref{app:additional}). Importantly, we find the same pattern generalizes across the other two models: $\pi_{0.5}$ peaks at over $35\%$ for a single-layer patch and over $70\%$ for a sliding patch, and MolmoAct2 at nearly $30\%$ and over $40\%$, respectively. In summary, in each model, a narrow band of only one to three adjacent layers accounts for most of the conditioning that shapes the action expert's behavior. We refer to this concentrated readout region as the \emph{action-conditioning interface}.

\section{Efficiency Improvements in Real Policies}
\label{sec:validation}

Our mechanistic findings suggest a direct opportunity to improve the efficiency of VLA architectures: the narrow VLM--action-expert interface identified in \Cref{fig:single_sliding_patch}. Thus, we restrict action expert conditioning to the identified layers by removing completely the cross attention layers outside of the \emph{action-conditioning interface}, and evaluate whether these simplified policies retain their capabilities.

\begin{table}[h]
\centering
\small
\setlength{\tabcolsep}{5pt}
\begin{tabular}{l c cc cc c}
\toprule
 & \textbf{Pick} & \multicolumn{2}{c}{\textbf{Pick \& Place}} & \multicolumn{2}{c}{\textbf{PnP Next-To}} & \textbf{Avg} \\
\cmidrule(lr){2-2}\cmidrule(lr){3-4}\cmidrule(lr){5-6}\cmidrule(lr){7-7}
Cross-attention layers & Success & Oracle & Final & Oracle & Final & (oracle) \\
\midrule
All 36 (released MolmoBot-DROID)      & \rB{62.0} & \rB{57.4} & \rB{49.6} & 21.2      & 15.1      & \rA{46.9 $\pm$ 1.7} \\
\addlinespace[2pt]
Layers 23--25, pretrained             & \rC{57.7} & \rA{59.2} & \rA{50.7} & \rB{23.3} & \rB{18.0} & \rB{46.7 $\pm$ 1.7} \\
Layer 24, pretrained                  & \rA{63.0} & 53.0      & 45.7      & 20.8      & 15.9      & \rC{45.6 $\pm$ 1.7} \\
\addlinespace[2pt]
Layers 23--25, adapt (50K FT)         & 56.8      & \rC{55.4} & \rC{45.9} & \rC{23.1} & \rC{17.8} & 45.1 $\pm$ 1.7 \\
Layer 24, adapt (50K FT)              & 54.7      & 54.0      & 44.9      & \rA{24.9} & \rA{19.4} & 44.5 $\pm$ 1.7 \\
\addlinespace[2pt]
Layer 12, adapt (50K FT)              & 25.7      & 20.9      & 16.3      & 6.9       & 4.9       & 17.8 $\pm$ 1.3 \\
\bottomrule
\vspace{0.1cm}
\end{tabular}
\caption{\textbf{Keeping only the action-conditioning interface preserves performance} Success rate (\%) on the MolmoBot paper's simulation benchmarks (1,000 identical episodes per benchmark). Reading only layer 24 (1 of 36 connections) comes within 1.3 points of the released all-layer model when trained from scratch. Reading only L23--25 comes within 0.2 points and gives the best Pick \& Place success of any model. $\pm$ is a 95\% binomial confidence interval. Final measures success state at end of episode. Oracle terminates at the first success state. Shading ranks each column (darkest = best).}
\label{tab:paper-protocol}
\end{table}

\subsection{Molmobot In Simulation}

We conduct our main evaluation with MolmoBot, whose openly available training data and simulation benchmarks support both training new policies and adapting an existing checkpoint \citep{deshpande2026molmobot}. Unlike the initial-timestep patching experiments, these evaluations test closed-loop behavior throughout complete manipulation tasks. Guided by the \emph{action-conditioning interface} located in layer-24 (\Cref{fig:single_sliding_patch}), we compare the released policy, which uses all 36 connections, against three interface variants: (i) policies trained with only the layer-24 interface, (ii) policies trained with only the layer-23-25 interface, and (iii) a policy trained with only the layer-12 in as a baseline control outside the \emph{action-conditioning interface}. For both (i) and (ii) we compare training the model from scratch and finetuning for 50k steps from the released checkpoint.

\Cref{tab:paper-protocol} reports results on three simulation benchmarks, with 1,000 episodes per benchmark and identical evaluation episodes across models. These are the provided MolmoBot benchmarks and use the MolmoSpaces simulator \cite{deshpande2026molmobot} for three tasks: Pick (pick up a target object), Pick and Place (pick up a target object and place it into a target location), and Next-To (pick up a target object and place it next to a named object). Overall, we observe that the policy trained with only the \emph{action-conditioning interface} is comparable to the original policy, with a 33\% reduction in FLOPS. 
The layer-24 pretrained policy achieves the highest Pick success (63.0\%), the layer-23-25 policy achieves the highest Pick \& Place oracle success (59.2\%), while the layer-24 adapted policy achieves the highest Next-To oracle success (24.9\%). Averaging Pick success and the two oracle metrics gives 46.9\% for the dense policy, 46.7\% for the layer-23-25 pretraining and 45.6\% for layer-24 pretraining. Thus, the limited connection interface supports almost identical overall performance. 

These architectures remove 33 to 35 of the 36 cross-attention modules ($92\%$-$97\%$ of the interface). Additionally, because nothing reads VLM layers 26--35, the VLM can stop early. Together, these cut the full policy's computation from 11.93 to 7.96 TFLOPs per action chunk, a $33\%$ reduction (\Cref{app:flops}). We see similar savings on real hardware. On a single H100 with a batch size of 64, producing one action chunk takes 209\,ms instead of 311\,ms, a $33\%$ latency reduction, and throughput rises from 21.5 to 30.9 chunks per second ($+44\%$). The throughput gain matters beyond single-robot control, for workloads that run many rollouts in parallel, such as large-scale simulation evaluation or reinforcement learning.

Selecting layers outside of the causally determined \emph{action-conditioning interface} hurts performance significantly. Under the same 50,000-update adaptation budget, retaining layer 12 yields only 25.7\% Pick success and 20.9\% Pick \& Place oracle success, compared with 54.7\% and 54.0\% when retaining layer 24. This contrast supports using the mechanistically identified readout region rather than selecting an arbitrary single-layer interface. Together, these results show that the localized action-conditioning pathway can guide both the design of a new policy and the simplification of an existing one with large efficiency gains, which we now show transfer to real-world deployment.

\subsection{$\pi_{0.5}$ in the real world}
We now test whether this design principle transfers to a different architecture in a physical embodiment. For $\pi_{0.5}$, we restrict action-expert attention to layers 14--16, masking this conditioning pathway at the remaining layers. We evaluate both policies on four tasks types for a total of 100 trials each. The four tasks types are: fruit in basket, [noun] in bin, marker in mug, and stack bowls.  As shown in \Cref{tab:pi05-real-robot}, the original and restricted-interface policies succeed in 64 and 67 out of 100 trails respectively. The efficiency gain is much smaller than for MolmoBot. Because $\pi_{0.5}$'s action expert reads prefix keys and values that the VLM has already computed, the ablation removes a relatively minor amount of flops. (\Cref{app:flops}). Averaged over the seven tasks, the median duration of successful trials nonetheless decreases from 34.9\,s to 26.5\,s, about 24\%. Because this does not arise from fast inference, it reflects a difference in how the restricted policy executes the task. Overall we conclude that there is no statistically significant different in task success rate between the two policies. However, the difference in task completion time likely reflect a difference in robustness between the policies. More discussion in \cref{app:pi05-real-robot}.

\begin{table}[h]
\centering
\small
\begin{tabular}{l c c}
\toprule
$\pi_{0.5}$ policy & Success & Task time (s) \\
\midrule
Original (all 18 layers)       & 64/100 (64\%) & 34.9 \\
Restricted (layers 14--16)     & 67/100 (67\%) & 26.5 \\
\bottomrule
\end{tabular}
\vspace{0.1cm}
\caption{\textbf{Real-robot evaluation of $\pi_{0.5}$.} Over 100 trials per policy on seven tabletop tasks
(\Cref{fig:pi05-real-tasks}; per-task results in \Cref{tab:pi05-real-robot-tasks}), restricting action-expert attention to layers 14--16 preserves task success. Task time is the median duration of successful trials for each task, averaged over the seven tasks. It measures full task execution, not inference latency.}
\label{tab:pi05-real-robot}
\end{table}

\section{Explaining the Action Conditioning Interface}
\label{sec:explaining}

\begin{figure}[t]
    \centering

    \begin{subfigure}{\linewidth}
        \centering
        \small
        \renewcommand{\arraystretch}{1.15}
        \begin{tabular}{@{}lccc@{}}
        \toprule
        Interchanged variable & \textbf{A} & \textbf{B}& \textbf{C}\\
        \midrule
        Visual $V$ & Change & Preserve & Change\\
        Semantic $S$ & Change & Change & Preserve\\
        Geometric $G$ & Preserve & Change & Change\\
        \bottomrule
        \end{tabular}
        \caption{Predicted causal signatures.}
        \label{tab:causal-signatures}
    \end{subfigure}

    \medskip

    \begin{subfigure}{\linewidth}
        \centering
        \includegraphics[width=\linewidth]{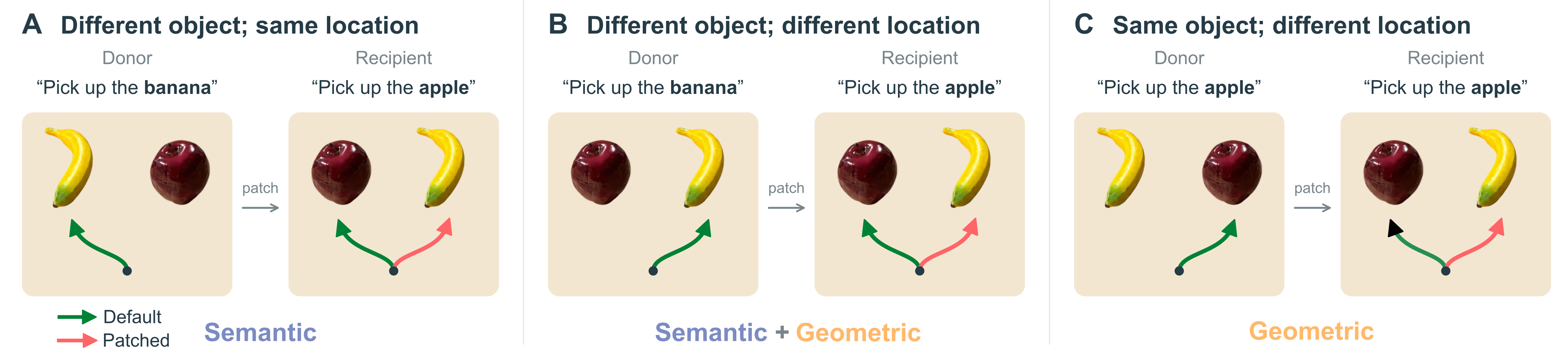}
        \caption{Predicted reaches under the labeled transfer hypotheses.}
        \label{fig:counterfactuals_predictions}
    \end{subfigure}

    \caption{
        \textbf{Three counterfactual comparisons test hypothesis about target selection} We patch activations from a donor run into a recipient run. (a) The input changes in \textbf{A}, \textbf{B} and \textbf{C} produce distinct causal signatures for each interchange intervention type. (b) Green arrows show the unpatched behavior; red arrows show predicted recipient behavior if the patch transfers the labeled information. These labels denote transfer \textit{hypotheses}, not information uniquely isolated by single comparisons. For example, in \textbf{A}, patching scene-layout information through an image patch can produce the same behavioral shift as patching target-object identity through a noun patch. We therefore interpret effects jointly across the comparisons and patched token positions.
    }
    \label{fig:counterfactuals}
\end{figure}

In a VLA, information flows from perception, through the vision-language backbone, to the action expert that produces motor commands. The interface discovery in Section~\ref{sec:methods} raises a question about what information the backbone supplies to the expert. Our goal is to distinguish representations that determine which object the model selects from those that determine where it reaches. We test hypotheses about this information by comparing the effects of interchange interventions with predictions from a high-level causal model. We use the intervention defined in \Cref{eq:interface_patch} to study the interface, and apply the same operation within the backbone to trace how that information forms upstream. %

\subsection{Causal variables and counterfactual predictions}
\label{sec:counterfactuals}

We model which of two locations the policy's actions target. For a fixed pair of objects, the visual variable $V$ maps each object to its location in the scene. The semantic variable $S$ identifies the object requested by the instruction. The geometric variable combines them to select the target location:
\begin{equation}
    G := V(S).
    \label{eq:geometric_variable}
\end{equation}
The output of the high-level model is the selected location. These variables specify hypotheses about the causal roles of neural representations. The full model, its variable domains, and its intervention predictions are defined in \Cref{app:causal_model}. %

Each hypothesis predicts how an interchange intervention changes the target. Interchanging $V$ locates the originally requested object in the counterfactual layout. An intervention on $S$ selects the counterfactually requested object at its location in the original scene. Interchanging $G$ directly transfers the counterfactual target location.

We construct two layouts containing the same two objects with their left--right arrangement exchanged, and pair each layout with instructions to pick up either object. Original--counterfactual pairs yield three comparisons, illustrated in \Cref{fig:counterfactuals}:
\begin{enumerate}[leftmargin=*, itemsep=3pt, parsep=0pt, topsep=3pt]
    \item \textbf{A: Different object, same target location.} Both the instruction and layout change, so the target objects occupy the same location. Interchanging $S$ or $V$ redirects the reach to the other location in the original scene, while interchanging $G$ alone preserves the original target location.

    \item \textbf{B: Different object, different target location.} The scene remains fixed while the instruction requests the other object. Interchanging either $S$ or $G$ redirects the reach toward the counterfactual target, so this comparison alone does not distinguish semantic from geometric information. The visual variable retains the same value.

    \item \textbf{C: Same object, different target location.} The instruction remains fixed while the layout changes. Interchanging $V$ or $G$ redirects the reach toward the counterfactual target location, which is occupied by the other object in the original scene. The semantic variable retains the same value, so interchanging $S$ alone preserves the original choice of object.
\end{enumerate}

Consider a scene which contains an apple on the left and a banana on the right, with the instruction to pick up the apple. In comparison \textbf{A}, the counterfactual scene swaps both the objects positions and the requested objects, meaning the geometric position of the target object remains the same. Interchanging the semantic variable $S$ makes the original run select the banana on its right, while an intervention on the geometric variable $G$ preserves the left target. The visual intervention also predicts a selection of the banana, because the originally requested apple is on the opposite side in the counterfactual layout. A noun intervention therefore redirects the reach by changing \emph{which} object the model selects, whereas an image intervention can redirect it by changing the visual location of the originally requested object. Comparison \textbf{B} separates the visual and semantic hypotheses, and comparison \textbf{C} completes the signatures in \Cref{tab:causal-signatures}.

The dataset contains 13 sets of object pairs for a total of 26 scenes (including object swaps). Each comparison uses 26 input pairs tested with both directions, for 52 ordered pairs per comparison and 156 across all three, \textbf{A}, \textbf{B}, and \textbf{C}. Repeating each ordered pair under eight noise seeds gives the 416 interventions per cell in the backbone results (\Cref{fig:vlm_backbone_patching}).

For each comparison, we vary the intervention positions $T$ at the interface and within the backbone, examining image tokens, the target noun, downstream prompt spans, and final readout positions. A intervention which redirects the robots action to the other object corresponds to $E=1$ and a preserving one to $E=0$. We first apply this test at the interface, then trace the variables upstream through the backbone.

\subsection{The action-conditioning interface is geometric}
\label{sec:interface_content}

We now have the necessary causal machinery to evaluate two hypothesis about the type of information exchanged in the action conditioning interface. We suggest 2 distinct hypothesis: 1. That the action conditioning interface only exchanges geometric information about the object positions and 2. that the action conditioning interface exchanges visual information about the scene. Each hypothesis predicts a different pattern across the three comparisons. The geometric hypothesis predicts a change in target in comparisons \textbf{B} and \textbf{C} and preservation of the target in \textbf{A} (\Cref{tab:causal-signatures}), while the visual hypothesis predicts a change in target in \textbf{A} and \textbf{C}, and the semantic hypothesis in \textbf{A} and \textbf{B}.

Our results on interface patching strongly support the geometric hypothesis. \Cref{fig:single_sliding_patch} averages comparisons \textbf{B} and \textbf{C}, in which the target location changes, and their effects concentrate in the narrow band of layers that defines the interface. Comparison \textbf{A}, which is not shown in the figure, changes both the requested object and the layout while keeping the target location fixed. \textbf{A} produces a shift of $|E|\leq0.022$ at every individual connection and three-layer window across all three models, including those where \textbf{B} and \textbf{C} peak. If the interface carried causally relevant information about the requested object or the scene layout, interventions in \textbf{A} would lead to a change in reach.

Because interface patches replace all token positions at a connection, a null effect in \textbf{A} alone could also be consistent with the interface carrying both visual and semantic variables and the action expert combining them into the geometric variable. The token-restricted backbone interventions in \Cref{sec:backbone_patching} rule this out. At the interface layers of each model patching the image tokens, the noun, or all text in comparison \textbf{A} changes $|E|\leq0.05$, whereas the same layers produce much larger effects (0.42--0.88) in comparisons \textbf{B} and \textbf{C}.

\subsection{How the backbone constructs the geometric variable}
\label{sec:backbone_patching}

We now turn to the VLM backbone and ask how scene information and the instruction combine into the geometric variable. To trace this construction, we apply the interchange operation from Eq.~\ref{eq:interface_patch} within the VLM. Let $z_{\ell,T}(x)$ denote residual-stream activations at layer $\ell$ and token positions $T$. We replace these activations with their counterfactual values:
\begin{equation}
    \bm a^{\mathrm{bb}}_{\ell,T}
    =
    \left.F(x^{\mathrm{o}})\right|_{
    z_{\ell,T}\leftarrow z_{\ell,T}(x^{\mathrm{cf}})}.
    \label{eq:backbone_patch}
\end{equation}
Activations at the remaining positions of layer $\ell$ retain their original values, while downstream backbone computation, interface conditioning, and action-expert computation proceed from the intervened state. This tests which representations affect behavior through their downstream effects.

\begin{figure}[t]
    \centering
    \includegraphics[width=\linewidth]{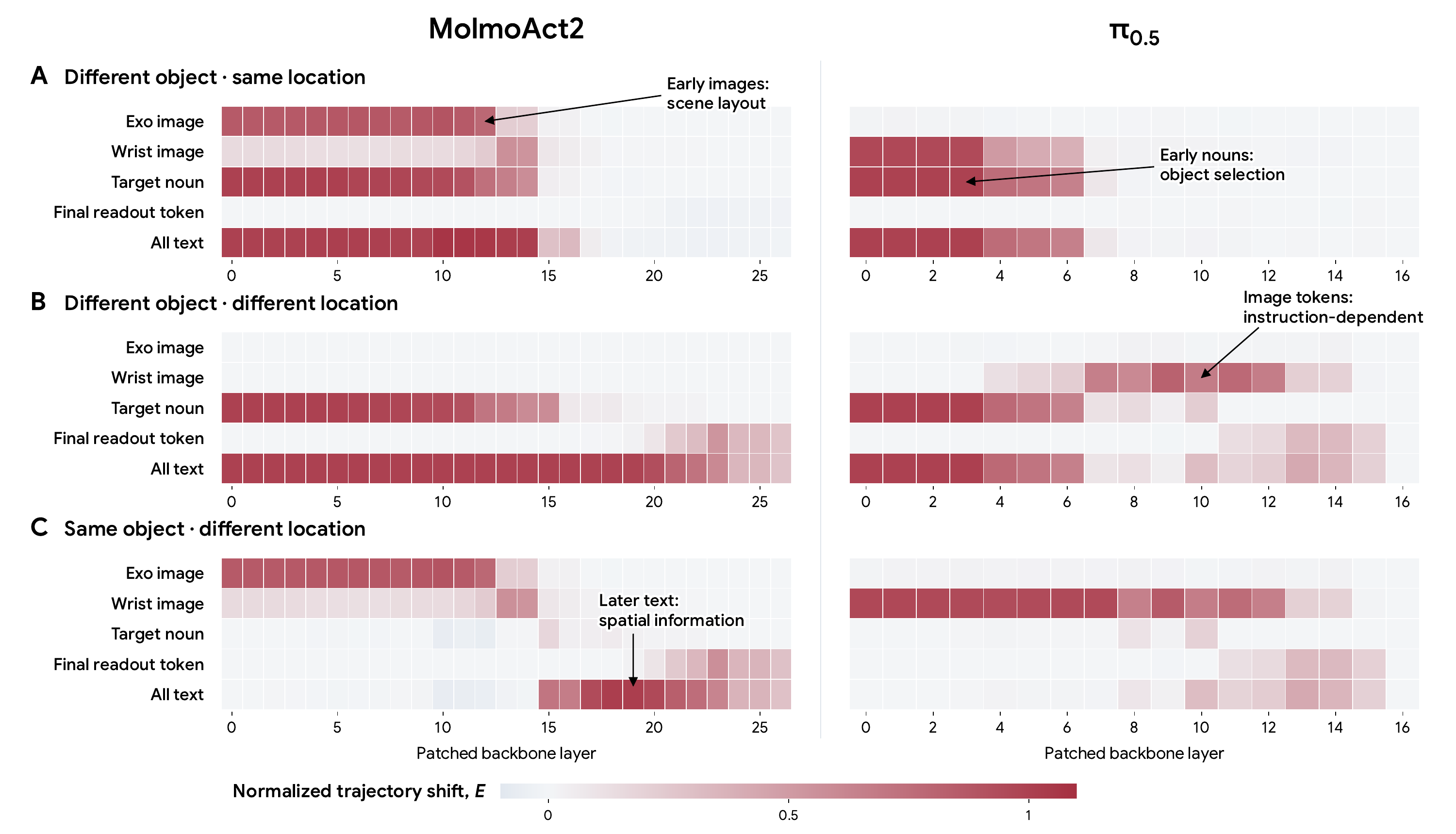}
    \caption{\textbf{Backbone interventions trace the construction of the geometric variable in MolmoAct2 and $\pi_{0.5}$.}
    Colors show normalized trajectory shift $E$ toward the other-object trajectory in the original scene.
    Each cell averages 416 interventions: 26 original--counterfactual pairs, each tested in both directions under eight noise seeds.
    The visual, semantic, and geometric hypotheses predict the signatures in \Cref{tab:causal-signatures}.
    We omit MolmoAct2 layers 27--35 as they have $|E|<0.001$.}
    \label{fig:vlm_backbone_patching}
\end{figure}

\Cref{fig:vlm_backbone_patching} shows backbone-intervention effects across token groups and layers for the three comparisons in \Cref{fig:counterfactuals}. Because each variable makes a distinct set of predictions across \textbf{A}, \textbf{B}, and \textbf{C} (\Cref{tab:causal-signatures}), we interpret each effect jointly from the comparison and the intervened positions. All experiments use the first timestep of each episode, when both objects are visible to both cameras.

\paragraph{Early image representations carry the visual variable.}
When the scene layouts differ (comparisons \textbf{A} and \textbf{C}), interchanging early image representations strongly redirects behavior. For the dominant camera, effects approach perfect reference recovery $E=1$ in early layers before declining at greater depth. The exterior view dominates in MolmoAct2, whereas the wrist view dominates in $\pi_{0.5}$. These effects are consistent with the visual hypothesis: transferring the counterfactual layout changes where the originally requested object is represented. In comparison \textbf{A}, an image intervention can therefore redirect the reach while leaving the original instruction unchanged.

\paragraph{Early noun representations carry the semantic variable.}
Interchanging early noun representations can instead change which object the model selects. Across all three models, these interventions can redirect the original run toward the counterfactually requested object. Comparison \textbf{A} distinguishes this semantic transfer from geometric transfer: the redirected reach follows that object's location in the original scene, even though the original and counterfactual targets occupy the same location. For example, transferring the representation of ``banana'' from a scene with the banana on the left can redirect an apple-instructed original run toward the banana on its right. The intervention therefore follows the prediction for $S$, selecting the counterfactually requested object at its original-scene location, an effect that cannot be explained by transferring the target location alone.

Early noun interventions also redirect behavior in comparison \textbf{B}, where the scene remains fixed and only the requested object changes. By contrast, noun interventions have little effect in comparison \textbf{C} in MolmoAct2, where the requested object remains fixed and its location changes. Together, this pattern matches the semantic signature, and the causal influence of the semantic variable at the noun position weakens with depth.

\paragraph{Later text representations carry the geometric variable.}
Comparison \textbf{C} holds the instruction fixed while changing the target object's location. Interchanging deeper text representations redirects the reach despite identical original and counterfactual text, supporting the presence of scene-dependent spatial information at these positions. At these layers, the pattern across \textbf{A}, \textbf{B}, and \textbf{C} follows the geometric signature: changes to the target location transfer through the text representations, while changes that preserve the target location have little effect. The backbone has therefore constructed a representation of which location to target.

In MolmoAct2, interchanging all text positions produces stronger effects than interchanging the final readout token alone. The readout token \texttt{<action\_output>} becomes an effective intervention site later, with geometric effects peaking around layers 22--23. These results indicate contributions beyond the readout token, alongside its increasing influence in deeper layers. A search over text positions identifies small groups that retain much of the effect of intervening on all text, with as few as four tokens retaining 89\% at layer 18 (\Cref{app:additional}). Together, these results trace how the geometric variable becomes available across text positions before the expert reads it at the interface. The high-level variable specifies only which location to target; its neural representation may be an explicit coordinate or a spatially specific motor plan that also supports further details of the movement.

\paragraph{Architecture shapes causal pathways within the model.}
Comparison \textbf{B} reveals an architectural difference in which positions can combine visual and semantic information. Here, original and counterfactual images are identical. In MolmoAct2 and MolmoBot, image positions cannot attend to the subsequent instruction, so their activations are unchanged by the instruction swap and interchanging them has no effect. In $\pi_{0.5}$, interchanging intermediate-layer image representations redirects behavior despite the identical visual inputs. Its bidirectional prefix attention allows image positions to incorporate the instruction.

Together with the image-intervention effects in comparison \textbf{C} and their decline in \textbf{A}, this result supports a geometric interpretation of $\pi_{0.5}$'s intermediate image representations. The geometric variable is therefore not restricted to positions associated with a particular input modality: text positions can carry the target location, and image positions can become conditioned on the requested task. \Cref{app:molmobot_backbone} reports the corresponding MolmoBot results. Across architectures, the backbone combines visual and semantic information into the geometric variable that is supplied through the action-conditioning interface. These interventions establish where each variable can be transferred, but do not show that it is represented exclusively at the identified positions.

\section{Discussion and conclusion}
\label{sec:discussion}
We used activation patching with controlled counterfactuals to trace how perception becomes action in three VLAs. Across MolmoBot, MolmoAct2, and $\pi_{0.5}$, the picture is consistent. 
Image tokens contain the scene layout in early layers and the target noun carries which object to pick. 
In deeper layers, the target's location emerges in text and readout position, as well as image positions in $\pi_{0.5}$, whose bidirectional attention lets them read the instruction.
The action expert then reads this information through the \emph{action-conditioning interface}, a narrow band of one to three middle-to-late layers. We then apply these findings to improve the efficiency of the policy. Keeping only the interface layers in MolmoBot cuts theoretical full-policy computation and measured latency by $33\%$, and raises batched throughput by $44\%$, while task success remains comparable to the dense policy. More broadly, these results suggest that mechanistic analysis can guide how robot policies are built, not only explain them after training.

\paragraph{Limitations.}
Our activation-patching experiments use the start of an episode where relevant objects are still in view. This is the only point where clean counterfactuals can be built because both objects are visible and the robot starts in the same state. Therefore our causal analysis may miss computation that matters in different phases of manipulation such as grasping. Furthermore our dataset only consists of 13 object-pairs in two binary locations. The closed-loop evaluations partly address these limitations. While the restricted policies were designed from start-of-episode interventions alone they still complete entire tasks, including grasping, placing and stacking, in scenes beyond the counterfactual dataset with comparable success rate over many trials (3000 for sim and 100 for real). If later phases of manipulation used fundamental different conditioning interfaces then we would expect lower success rates.

\newpage
\bibliography{VLA_Patching_main}
\bibliographystyle{bibliostyle}

\newpage
\appendix
\clearpage
\section{A Causal Model of Object Selection and Target Location}
\label{app:causal_model}

This appendix defines the high-level causal model used in \Cref{sec:counterfactuals} \citep{geiger2021causal,geiger2024causalabstractiontheoreticalfoundation}. The model we propose describes pick tasks with three variables: the scene layout, the requested object, and the target location. For each interchange intervention, it predicts which of the two locations the will policy reach toward. These predictions give comparisons \textbf{A}, \textbf{B}, and \textbf{C} the same interpretation across models and intervention sites.

\paragraph{Inputs and variables.}
Fix a pair of objects $\mathcal O=\{a,b\}$ and two target locations $\mathcal P=\{p_{\mathrm L},p_{\mathrm R}\}$. A layout places each object at one of the two locations, thus it is a one-to-one map from objects to locations:
\begin{equation}
\mathcal V=\operatorname{Bij}(\mathcal O,\mathcal P).
\end{equation}
There are two possible layouts. The model takes a layout $L\in\mathcal V$ and a requested object $Q\in\mathcal O$ as inputs. The corresponding policy input is $x=(I(L),u(Q),s)$, where $I(L)$ are the camera images of the layout, $u(Q)$ is the instruction requesting $Q$, and $s$ is the robot state. Crossing the two layouts with the two requests gives four inputs per object pair. The robot state is the same within each comparison, and we share the generation noise $\epsilon$.

The model has four internal variables. The visual variable $V\in\mathcal V$ maps each object to its location. The semantic variable $S\in\mathcal O$ identifies the requested object. The geometric variable $G\in\mathcal P$ is the location of that object, and the output $Y\in\mathcal P$ is the location the actions target.

\paragraph{Structural equations.}
The causal model $\mathcal M$ is defined by
\begin{equation}
\label{eq:highlevel-structural-equations}
V=L,\qquad S=Q,\qquad G=V(S),\qquad Y=G.
\end{equation}
The scene layout $L$ and the requested object $Q$ together determine the target location, $G=V(S)$. The output $Y$ is action that the robot policy produces and it takes the values of $G$ through and edge from $G$ to $Y$. The model only covers which of the two object locations the policy reaches toward. Describing other characteristics such as the path, grasp or speed would require additional variables.

\paragraph{Interchange interventions.}
Let $x^{\mathrm{o}}$ be the original input and $x^{\mathrm{cf}}$ the counterfactual input, and write $Z^{\mathrm{o}}$ and $Z^{\mathrm{cf}}$ for the value a variable $Z$ takes when $\mathcal M$ runs on each. An interchange intervention on $Z\in\{V,S,G\}$ runs the model on the original input but sets $Z$ to its counterfactual value.
\begin{equation}
\label{eq:highlevel-interchange}
Y^{Z}(x^{\mathrm{o}},x^{\mathrm{cf}})
=Y_{\mathcal M}\!\left(x^{\mathrm{o}};
\operatorname{do}(Z\leftarrow Z^{\mathrm{cf}})\right).
\end{equation}
All other variables are computed as usual from the original input and the intervened value. Substituting into \Cref{eq:highlevel-structural-equations} gives
\begin{align}
Y^{V}(x^{\mathrm{o}},x^{\mathrm{cf}})&=V^{\mathrm{cf}}(S^{\mathrm{o}}),\\
Y^{S}(x^{\mathrm{o}},x^{\mathrm{cf}})&=V^{\mathrm{o}}(S^{\mathrm{cf}}),\\
Y^{G}(x^{\mathrm{o}},x^{\mathrm{cf}})&=V^{\mathrm{cf}}(S^{\mathrm{cf}}).
\label{eq:highlevel-predictions}
\end{align}
Put simply, interchanging $V$, the visual variable, produces an action toward where the originally requested object sits in the counterfactual layout. Interchanging $S$, the semantic variable, sends it to the counterfactually requested object in the original layout. Interchanging $G$, the geometric variable copies the counterfactual target location directly.

\paragraph{Predictions for comparisons A, B, and C.}
Let layout $V_0$ place $a$ at $p_{\mathrm L}$ and $b$ at $p_{\mathrm R}$, and let $V_1$ swap them. Take the original input to be layout $V_0$ with a request for $a$, so the original run reaches to $p_{\mathrm L}$. \Cref{tab:highlevel-predictions} lists the counterfactual input for each comparison and the location each intervention predicts. Relabeling the objects or the locations gives the same table.

\begin{table}[t]
\centering
\small
\begin{tabular}{lccc}
\toprule
 & A & B & C \\
\midrule
Counterfactual layout & $V_1$ & $V_0$ & $V_1$ \\
Counterfactual request & $b$ & $b$ & $a$ \\
Clean original output & $p_{\mathrm L}$ & $p_{\mathrm L}$ & $p_{\mathrm L}$ \\
Clean counterfactual output & $p_{\mathrm L}$ & $p_{\mathrm R}$ & $p_{\mathrm R}$ \\
\midrule
Visual intervention $Y^{V}$ & $p_{\mathrm R}$ & $p_{\mathrm L}$ & $p_{\mathrm R}$ \\
Semantic intervention $Y^{S}$ & $p_{\mathrm R}$ & $p_{\mathrm R}$ & $p_{\mathrm L}$ \\
Geometric intervention $Y^{G}$ & $p_{\mathrm L}$ & $p_{\mathrm R}$ & $p_{\mathrm R}$ \\
\bottomrule
\end{tabular}
\caption{\textbf{High-level predictions for the original input $(V_0,a)$.} Each entry is the location the actions target. $p_{\mathrm L}$ preserves the original target ($E=0$), and $p_{\mathrm R}$ redirects the reach to the other object's location in the original scene ($E=1$). The intervention rows give the signatures in
\Cref{tab:causal-signatures}.}
\label{tab:highlevel-predictions}
\end{table}

Each variable ($V$, $S$, $G$) preserves the original target in exactly one comparison (\textbf{A}, \textbf{B}, \textbf{C}). The visual variable is isolated in \textbf{B}, where the layout is constant; the semantic variable in \textbf{C}, where the instruction is constant; and the geometric variable in \textbf{A}, where the target location is constant. Using these patterns we can determine which variable a given intervention carries by running interchange interventions across all three comparisons. We therefore use the same token positions and intervention type across \textbf{A}, \textbf{B}, and \textbf{C}.

\paragraph{Relation to neural interventions.}
The neural counterparts are the interface intervention $\bm a^{\mathrm{int}}_{J,T}$ (\Cref{eq:interface_patch}), which replaces the conditioning supplied to expert blocks $J$ while the original backbone computation stays fixed, and the backbone intervention $\bm a^{\mathrm{bb}}_{\ell,T}$ (\Cref{eq:backbone_patch}), which replaces residual-stream activations at layer $\ell$ and positions $T$ and lets all downstream computation proceed from the intervened state. The hypothesis that a site carries variable $Z$ predicts that intervening there sends the reach to the location given by $Y^Z$. Let $\rho_Y$ read off which of the two locations an action sequence targets. Exact agreement at a backbone site would require
\begin{equation}
\label{eq:highlevel-neural-agreement}
\rho_Y\!\left(\bm a^{\mathrm{bb}}_{\ell,T}(x^{\mathrm{o}},x^{\mathrm{cf}})\right)
=Y^Z(x^{\mathrm{o}},x^{\mathrm{cf}})
\end{equation}
for every tested input pair, and the same condition with $\bm a^{\mathrm{int}}_{J,T}$ applies to an interface site. The normalized trajectory shift (\Cref{eq:behavioral_transfer}) measures how closely the policy meets this criterion: a predicted redirection to $p_{\mathrm R}$ corresponds to $E=1$, and a predicted preservation of $p_{\mathrm L}$ to $E=0$.

\paragraph{The handover to action.}
The geometric variable $G$ is the information we propose the backbone passes to the action expert. We call the point where a representation of $G$ is supplied to the expert the handover to action. The interface interventions locate this point in each model, and the high-level model predicts its effect on behavior through the edge $G\to Y$.

\section{Additional Results}
\label{app:additional}

\subsection{Action-conditioning interface}
\label{app:full_trajectory}

\begin{figure}[h]
    \centering
    \begin{subfigure}[t]{0.49\linewidth}
        \centering
        \includegraphics[width=\linewidth]{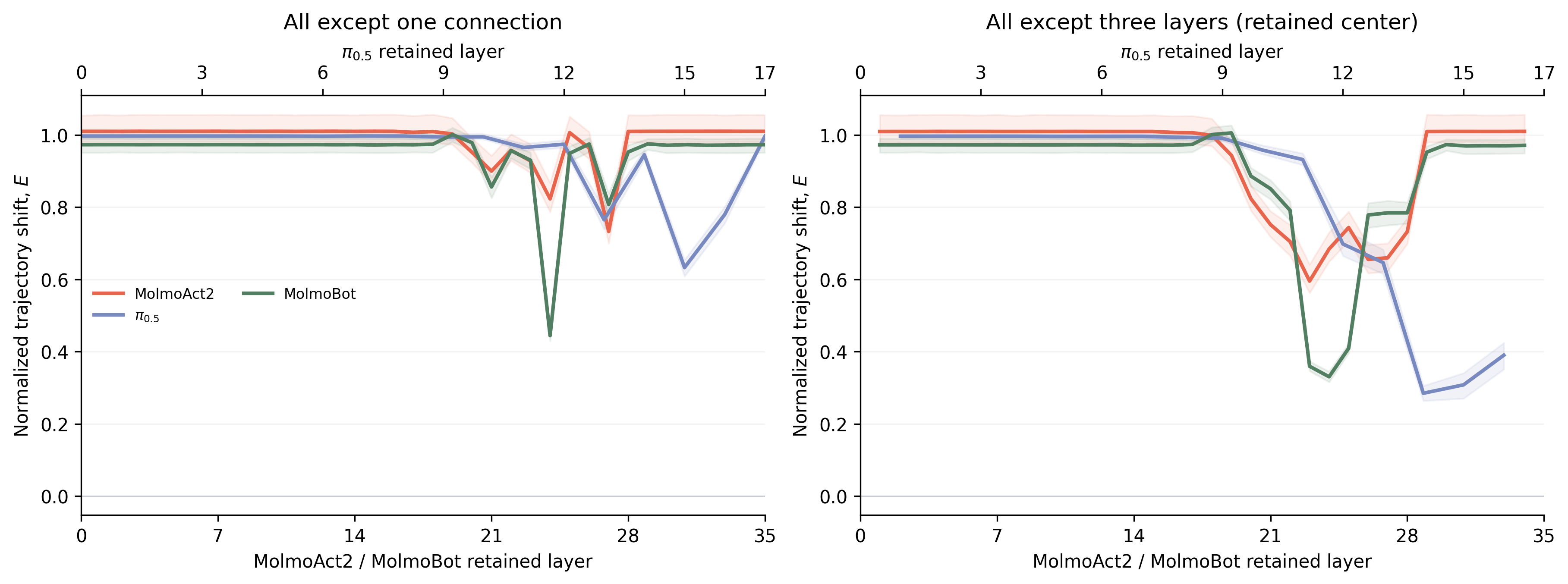}
        \caption{Patching all connections but selected}
        \label{fig:interface_complement}
    \end{subfigure}\hfill
    \begin{subfigure}[t]{0.49\linewidth}
        \centering
        \includegraphics[width=\linewidth]{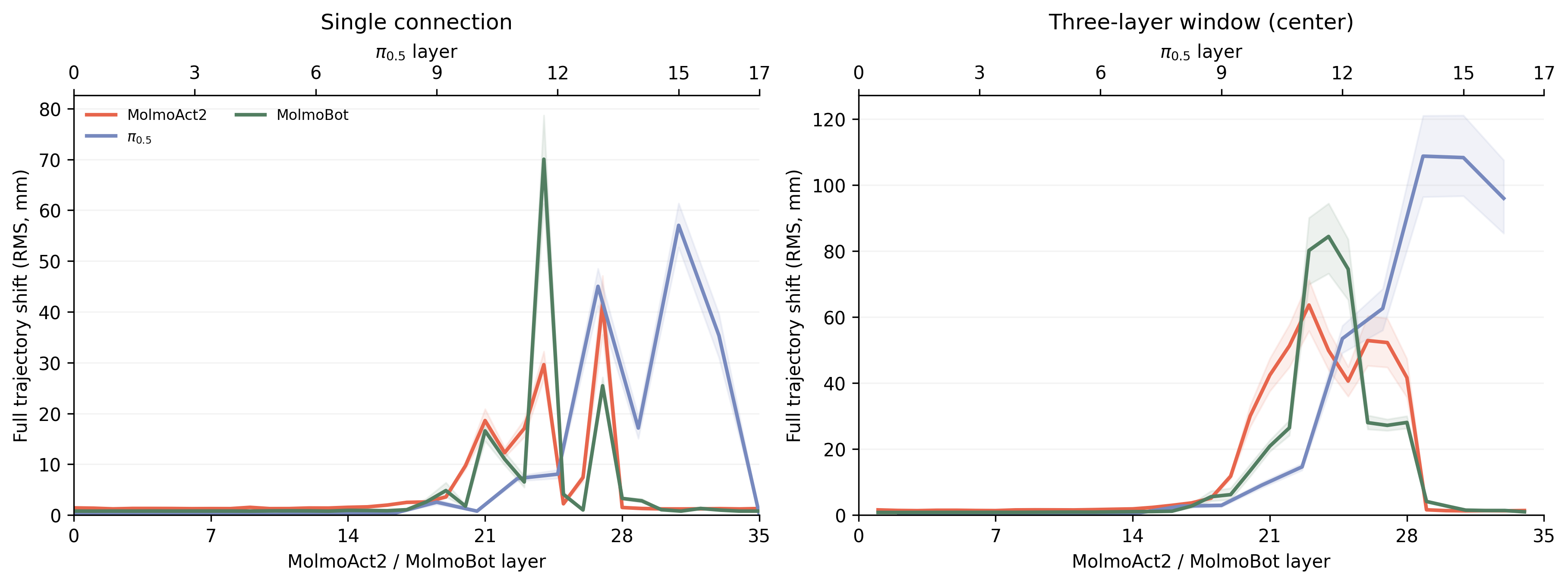}
        \caption{Interface patching with RMS rather than $E$}
        \label{fig:interface_full_trajectory}
    \end{subfigure}
    \caption{\textbf{Complementary and full-trajectory interface patching.}
    \textbf{(a)} Normalized trajectory shift $E$ when patching all connections except one (left) or except three consecutive connections (right). Horizontal axes index the unchanged connection or window center.
    \textbf{(b)} RMS displacement in millimeters after patching only a single connection (left) or three consecutive connections (right, indexed by their center). Trajectories are reconstructed from predicted actions using forward kinematics.
    Both panels average comparisons B and C with equal weighting across object-pair families; shading denotes 95\% family-bootstrap confidence intervals.}
    \label{fig:interface_complement_full}
\end{figure}

We test the joint influence of the remaining connections by patching every connection except one or a three-layer window, leaving the original conditioning at the excluded connections unchanged. Figure~\ref{fig:interface_complement} shows that preserving the strongest connections substantially reduces behavioral transfer, but appreciable effects remain. These results support concentrated influence, but not completely negligible joint influence outside the strongest one to three connections.

Because $E$ measures displacement along a reference contrast, it can miss orthogonal changes. We therefore also evaluate the original single-connection and three-layer interventions using the RMS displacement between patched and original end-effector trajectories, $R$, computed before averaging interventions. The dominant peaks remain unchanged across all three models (Figure~\ref{fig:interface_full_trajectory}), showing that the observed concentration persists when measuring changes in all Cartesian directions.

\subsection{Backbone Patching in MolmoBot}
\label{app:molmobot_backbone}

\begin{figure}[h]
    \centering
    \includegraphics[width=\linewidth]{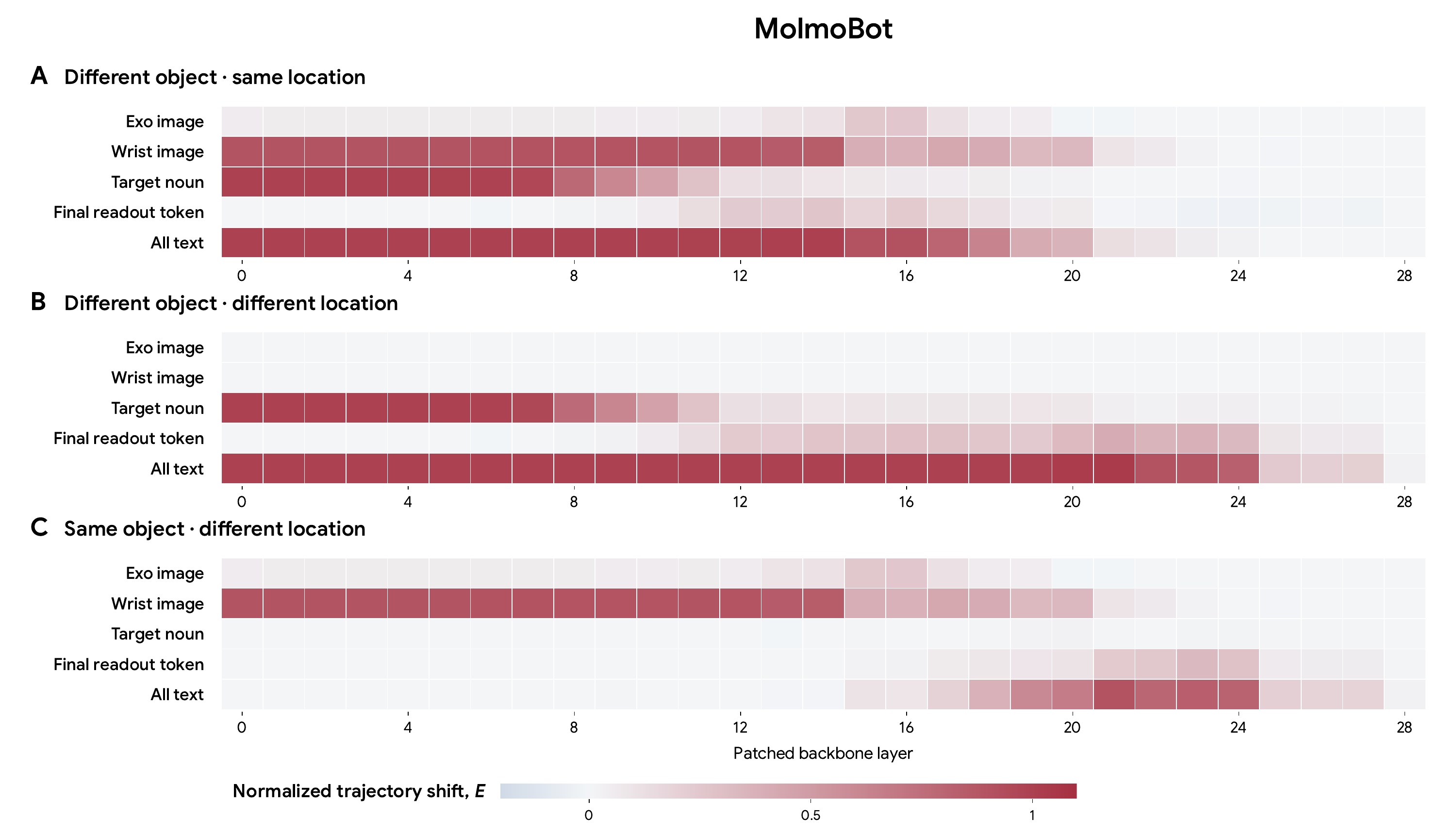}
    \caption{\textbf{MolmoBot backbone patching across 13 object pairs.} Same comparisons, token groups, and metric as \Cref{fig:vlm_backbone_patching}. Each cell averages 416 patches: 26 donor--recipient pairs, each patched in both directions under eight noise seeds. Image patches in B are exactly zero because MolmoBot's image positions cannot attend to the instruction. Omitted layers 29--35 have $|E|<0.005$.}
    \label{fig:app_molmobot_abc}
\end{figure}

\Cref{fig:app_molmobot_abc} shows backbone patching for MolmoBot on the same 13 object pairs as \Cref{fig:vlm_backbone_patching}. We see that its causal structure closely matches MolmoAct2's. In both models, the target noun carries object identity in early layers (A and B). Location is then distributed later across text positions (C). In both MolmoACT2 and MolmoBot, image patches have no effect in B, because image positions precede the instruction and cannot attend to it. The main difference we see visually separating the two models is which camera matters: the wrist view dominates in MolmoBot, and the exterior view dominates in MolmoAct2.

This similarity between the models causal structure holds even though the two models are trained on completely different data (simulated versus real demonstrations) and their action experts attend to the backbone differently. In $\pi_{0.5}$, whose backbone attends bidirectionally over the prefix image tokens absorb some instruction conditioned information and causal influence (\Cref{fig:vlm_backbone_patching}). Together, the three models suggest that the backbone architecture, and in particular its attention mask, shapes where task information lives in a VLA.  

\subsection{Text Tokens That Store Geometric Location in MolmoAct2}
\label{app:minimal_sets}

\begin{figure}[h]
    \centering
    \includegraphics[width=\linewidth]{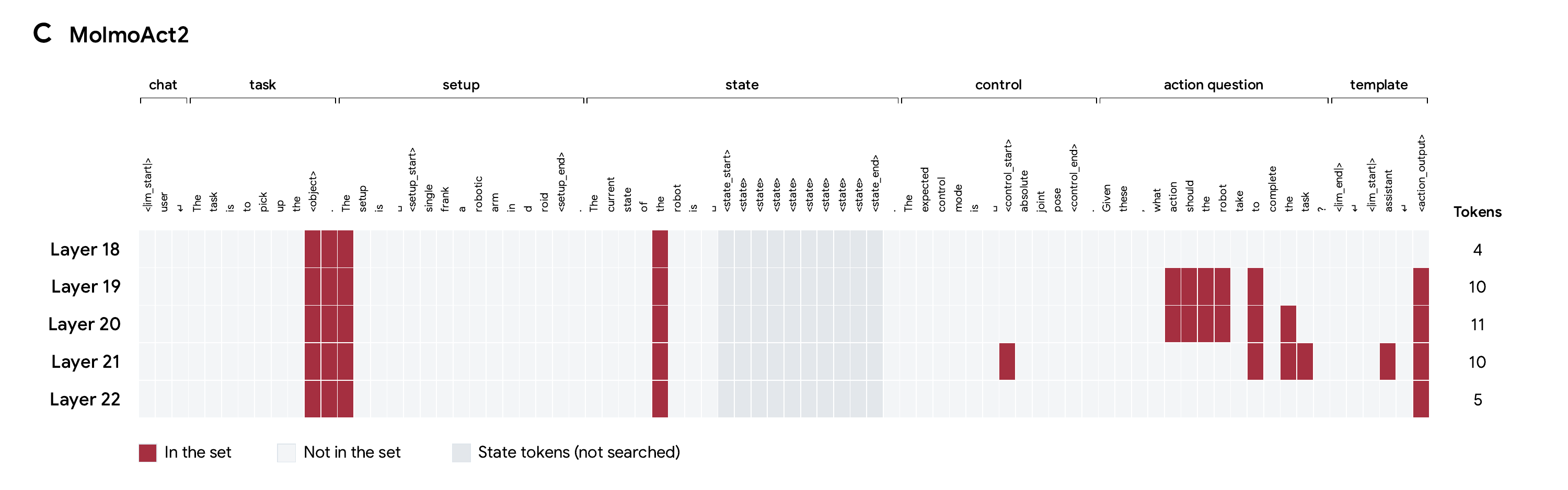}
    \caption{\textbf{Small token sets that carry location in MolmoAct2, across all 13 object pairs.} For each layer, filled cells mark the smallest set of text tokens, found by greedy forward selection, whose joint patch keeps at least $75\%$ of the effect of patching all 79 non-state text tokens in comparison C; the right column gives the number of tokens. Effects are averaged over all 52 comparison-C donor--recipient pairs with equal weight per object pair. The sets keep $76$--$89\%$ of the effect on the noise seed used for selection and $72$--$90\%$ on three held-out seeds. The target noun, which differs across object pairs, is labeled \texttt{<object>}; state tokens were not searched.}
    \label{fig:app_minimal_sets}
\end{figure}

Based on the backbone patching results, particularity with respect to geometry, we know that the policy represents information about the target object position in mid-to-late layer text tokens. Geometric information is distributed across many different tokens. To identify which text positions carry this location, we search for a small set of tokens which jointly patch recovers the effect of patching all of the text. The search runs over all 52 comparison C pairs and the results for MolmoAct2 are shown in \Cref{fig:app_minimal_sets}. 

At layer 18, four tokens suffice: the target noun, the period that ends the task sentence, the ``The'' that follows it, and the ``the'' in ``the current state of the robot''. Together they keep $89\%$ of the effect. At layers 19--21 the sets grow to 10--11 tokens that add \texttt{<action\_output>} and words of the action question (and, at layer 21, \texttt{<control\_start>} and \texttt{assistant}), keeping $76$--$77\%$. At layer 22 the action question drops out, and \texttt{<action\_output>} with the four sentence-boundary tokens keeps $77\%$. Removing any single token from any set drops retention below $75\%$. Location is therefore carried jointly by a few structural positions (sentence boundaries, function words, and the action question) before it concentrates on the readout token. We hypothesize this is where the computation of the target object location is happening.

\section{Compute and Latency}
\label{app:flops}
Here we provide more details on our theoretical FLOP calculations and empirical latency and throughput results for MolmoBot. One MolmoBot action chunk consists of one vision-encoder pass, one VLM pass, and ten action-expert denoising passes. Restructuring the cross-attention blocks also allows us to terminate the VLM computation early as nothing reads VLM layers 25--35.

\paragraph{Estimated operation counts.}
We assume that a dense layer with $N$ parameters applied to $T$ tokens costs about $2NT$ FLOPs, adding attention products and ignoring normalization, softmax, and embeddings. A VLM layer which for Molmobot is $\approx$101M parameters and 928 tokens costs 0.20 TFLOPs. Therefore running only 25 out of 36 layers reduces the VLM computation from 7.25 to 5.04 TFLOPs. A cross-attention block costs 5.9 GFLOPs per flow step, the majority of it from projecting the 928 context tokens into the expert. 36 blocks over ten steps cost 2.13 TFLOPs while the rest of the action expert only costs 0.08 TFLOPs. Therefore keeping one block reduces the expert from 2.22 to 0.14 TFLOPs. With the vision encoder fixed at 2.46 TFLOPs, the full policy falls from 11.93 to 7.64 TFLOPs per action chunk overall. We apply the same calculations to the 36 layer policy and L23-25 policy and show the results in \Cref{tab:flops}. We note that the released MolmoBot implementation does not cache the cross-attention context projection and keys/values across the ten denoising passes. Doing so would slightly reduce the cost of every policy.

For $\pi_{0.5}$ the saving is small because its action expert reads prefix keys and values that the VLM has already computed. Most of the gain comes from letting the VLM stop after layer 16 (0.23 TFLOPs). The full policy falls from 4.76 to 4.52 TFLOPs ($-5\%$). Given how small these estimated saving are, we do not empirically analyze $\pi_{0.5}$'s latency or throughput.

\begin{table}[h]
\centering
\small
\begin{tabular}{l rrr r r}
\toprule
 & Vision & VLM & Action expert & Total & Change \\
\midrule
MolmoBot, all 36 layers   & 2.46 & 7.25 & 2.22 & 11.93 & -- \\
MolmoBot, layer 24        & 2.46 & 5.04 & 0.14 & 7.64  & $-36\%$ \\
MolmoBot, layers 23--25   & 2.46 & 5.24 & 0.26 & 7.96  & $-33\%$ \\
\addlinespace[2pt]
$\pi_{0.5}$, all 18 layers & 0.66 & 3.98 & 0.13 & 4.76 & -- \\
$\pi_{0.5}$, layers 14--16 & 0.66 & 3.75 & 0.11 & 4.52 & $-5\%$ \\
\bottomrule
\end{tabular}
\caption{Estimated TFLOPs per action chunk. Restricted policies stop the VLM after the deepest layer the action expert reads.}
\label{tab:flops}
\end{table}

\paragraph{Measured latency and throughput.}
We timed MolmoBot on a single H100 in bfloat16 with ten flow-matching steps and two cameras $\times$ two frames (\Cref{tab:timing}). At batch size 1, the saving comes almost entirely from the action expert. At batch size 64, the all-layer expert's 36 cross-attention blocks become compute-bound, so the throughput gain is larger. The measured latency reduction ($35\%$) matches the estimated full-policy reduction ($36\%$).

\begin{table}[h]
\centering
\small
\begin{tabular}{l c c c}
\toprule
MolmoBot & Latency (ms) & Throughput (chunks/s) & Memory (GB) \\
\midrule
All 36 layers  & $311\pm3$            & 21.5           & 42.7 \\
Layer 24       & $204\pm16$ ($-35\%$) & 31.8 ($+48\%$) & 31.7 \\
Layers 23--25  & $209\pm2$ ($-33\%$)  & 30.9 ($+44\%$) & 32.2 \\
\bottomrule
\end{tabular}
\caption{Measured inference cost on one H100 (bfloat16, ten flow-matching steps). Latency is at batch size 1; throughput and peak memory are at batch size 64. Restricted policies stop the VLM after the deepest layer the action expert reads.}
\label{tab:timing}
\end{table}

\section{Implementation Details}
\label{app:implementation}
\paragraph{MolmoBot evaluation.}
All MolmoBot results use the evaluation code released with MolmoBot~\citep{deshpande2026molmobot}. We evaluate on three fixed sets of 1,000 episodes (Pick, Pick \& Place, Pick \& Place Next-To) in held-out ProcTHOR-Objaverse houses, run in the MolmoSpaces simulator with the released configuration. Every policy is evaluated the same set of episodes with identical noise seeds. 
\paragraph{Model Training.}
We use the released MolmoBot training code with its two-stage recipe, except modifying the set of action-expert blocks that keep cross-attention.  Stage~1 initialises from Molmo2-4B and trains for 200K steps on single-frame observations then stage~2 continues for another 50K steps with two of horizon frames. The \emph{adapt} variants start from the released MolmoBot-DROID checkpoint and then run the stage-2 recipe for 50K steps.  For $\pi_{0.5}$ both policies are fine-tuned from the released $\pi_{0.5}$ base checkpoint~\citep{intelligence2025pi05} on DROID~\citep{khazatsky2024droid} with the published openpi DROID recipe.

\subsection{Experimental Setup}
\label{app:pi05-real-robot}

\begin{figure}[h]
    \centering
    \includegraphics[width=\linewidth]{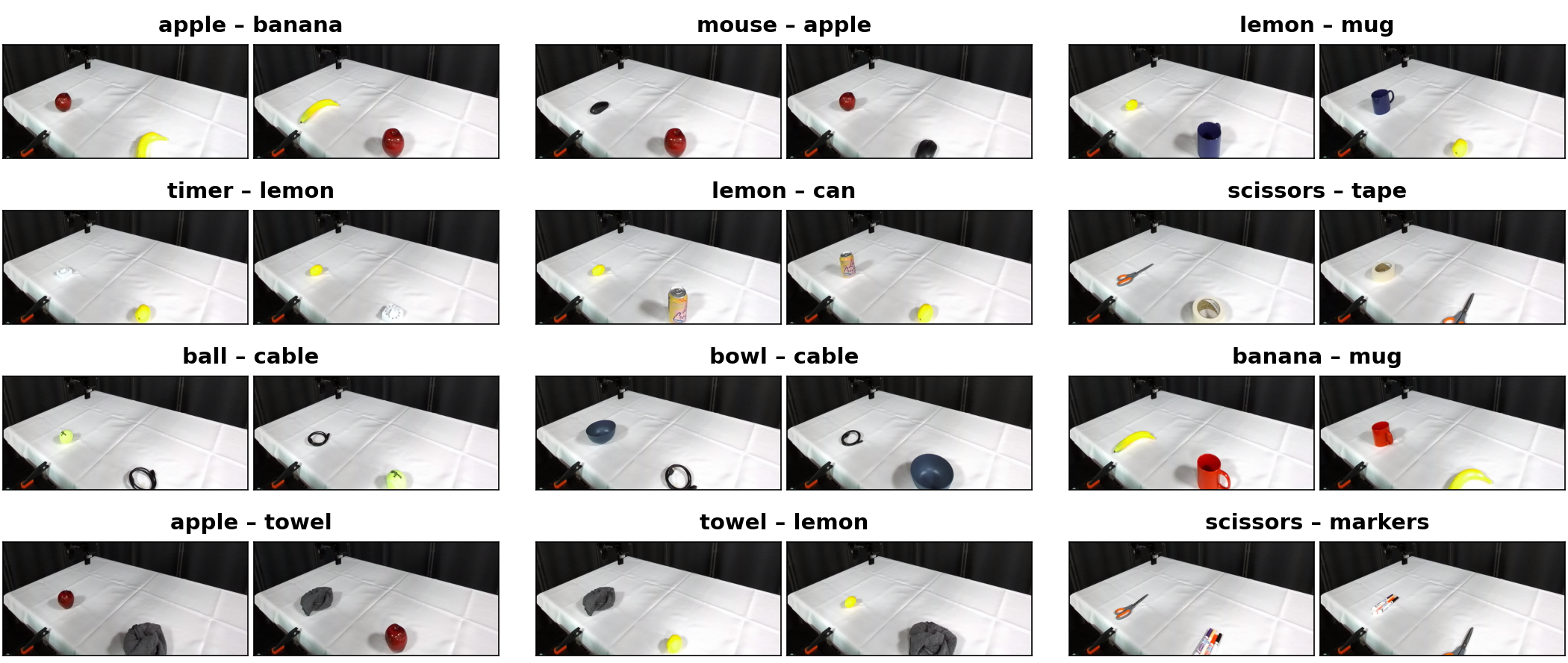}
    \caption{\textbf{Counterfactual scenes.} Object pairs used each shown in both layouts with the objects' left--right positions exchanged. Images show the exterior DROID camera, cropped to the table (during experiments the full frame is used).}
    \label{fig:app_scenes}
\end{figure}

\paragraph{Counterfactual dataset creation.}
We collect data on 13 pairs of household objects on a DROID setup in two layouts to form (26 scenes; \Cref{fig:app_scenes}). We pair each scene with an instruction to pick up either object, giving a total of 52 inputs.

\begin{figure}[h]
\centering
\includegraphics[width=\linewidth]{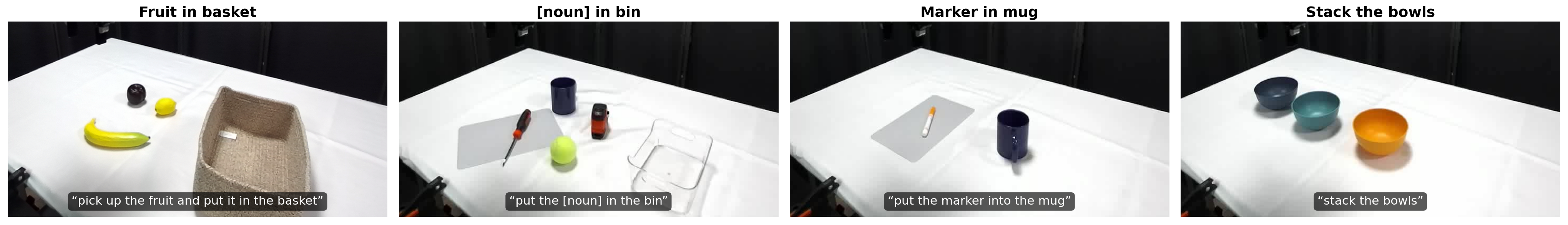}
\caption{\textbf{Real-robot tasks for $\pi_{0.5}$.} The first exterior-camera frame of each task, with the instruction given to both policies. For the [noun] in bin task,[noun] is replaced with each of the target objects for 10 trials each.}
\label{fig:pi05-real-tasks}
\end{figure}

\paragraph{Real world evaluations}
We evaluate the original and restricted $\pi_{0.5}$ policies on four tabletop tasks (\Cref{fig:pi05-real-tasks}), with 100 trials per policy. The fruit-in-basket task places three pieces of fruit into a basket. In the next task, the robot places one of four objects (a screwdriver, a mug, a tape measure or a tennis ball) in a clear bin to test language following abilities. The remaining tasks place a marker in a mug and stack three bowls, mean to test general dexterity. We do not find a statistically significant difference in success rate between the two policies (67 vs.\ 64 of 100 trials; Fisher's exact test, $p=0.77$). However, the restricted policy completes trials faster. Averaged over the seven tasks, the median duration of successful trials falls from 34.9 to 26.5\,s (24\% shorter; permutation test with policy labels shuffled within each task, $p=0.01$). Due to small sample sizes no single task reaches significance on its own. We conjecture that the representations attended to in L14-16 may be less prone to overfitting and therefore after fine-tuning the $\pi_{0.5}$ model may have more robust representations.

\begin{table}[h]
\centering
\small
\begin{tabular}{l c cc cc}
\toprule
 & & \multicolumn{2}{c}{Original (all 18 layers)} & \multicolumn{2}{c}{Restricted (layers 14--16)} \\
\cmidrule(lr){3-4} \cmidrule(lr){5-6}
Task & Trials & Success & Time (s) & Success & Time (s) \\
\midrule
Fruit in basket        &  25 & 22 &  53.6 & 22 &  45.4 \\
Marker in mug          &  20 & 15 &  16.1 & 10 &  16.4 \\
Stack the bowls        &  15 &  8 &  71.8 & 11 &  50.0 \\
Screwdriver in bin     &  10 &  5 &  37.8 &  7 &  20.0 \\
Mug in bin             &  10 &  3 &  18.9 &  8 &  21.4 \\
Tape measure in bin    &  10 &  3 &  30.7 &  2 &  20.8 \\
Tennis ball in bin     &  10 &  8 &  15.3 &  7 &  11.5 \\
\midrule
All tasks              & 100 & 64 &  34.9 & 67 &  26.5 \\
\bottomrule
\end{tabular}
\caption{\textbf{Real-robot results for $\pi_{0.5}$ by task.} Success counts and median duration of successful trials. The last row gives total successes and the median task time averaged over the seven tasks.}
\label{tab:pi05-real-robot-tasks}
\end{table}

\end{document}